\documentclass[runningheads]{llncs}
\usepackage[pdftex]{graphicx}
\usepackage{comment}
\usepackage{amsmath,amssymb} 
\usepackage{color}
\usepackage{xcolor}

\usepackage[width=122mm,left=12mm,paperwidth=146mm,height=193mm,top=12mm,paperheight=217mm]{geometry}

\begin{document}
\pagestyle{headings}
\mainmatter
\def\ECCVSubNumber{4967}  

\title{Can a powerful neural network be a teacher\\ for a weaker neural network?} 

\author{
    Nicola Landro,Ignazio Gallo,Riccardo La Grassa, \\
    University of Insubria \\
    Varese, Italy\\
    e-mails: \{nlandro, ignazio.gallo, rlagrassa\}@uninsubria.it
}

\maketitle

\begin{abstract}
The transfer learning technique is widely used to learning in one context and applying it to another, i.e. the capacity to apply acquired knowledge and skills to new situations.
But is it possible to transfer the learning from a deep neural network to a weaker neural network? 
Is it possible to improve the performance of a weak neural network using the knowledge acquired by a more powerful neural network? 
In this work, during the training process of a weak network, we add a loss function that minimizes the distance between the features previously learned from a strong neural network with the features that the weak network must try to learn.
To demonstrate the effectiveness and robustness of our approach, we conducted a large number of experiments using three known datasets and demonstrated that a weak neural network can increase its performance if its learning process is driven by a more powerful neural network.
\keywords{deep neural network, transfer learning, image classification}
\end{abstract}

\section{Introduction}
\begin{figure}
    \centering
    \includegraphics[width=.45\textwidth]{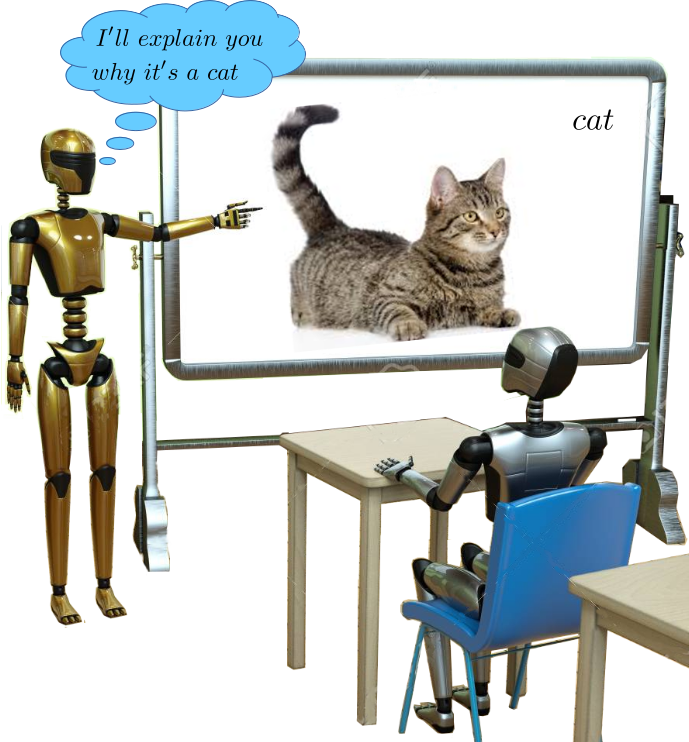}
    \caption{The main idea is to use the acquired knowledge of a very powerful neural network and transfer it to a simpler and weaker neural network. In the same way, as a teacher transfers his knowledge to a pupil trying to explain to him why.}
    \label{fig:basic_idea}
\end{figure}

For a pupil having a good teacher is certainly the guarantee to be able to learn a certain subject better.
A pupil who learns a new topic without a good teacher cannot reach the same level of knowledge as another pupil with identical potential but who is followed by a good teacher.
A teacher does not simply tell you what to study but also explains how best to organize the various concepts to better assimilate and understand them.
As shown in the example of Fig.~\ref{fig:basic_idea}, a student can independently understand what a cat is by analyzing its image. But if we want the student to become an expert in cats then the contribution of an expert teacher will also be needed to explain to him what the characteristic elements of the cat are, for example, the cat has a tail, a moustache, a coat, etc.
In this work, we want to transfer the same reasoning made for a teacher and his pupil, on the training process for a deep neural network.
So we propose a model to improve the accuracy of a weaker model transferring knowledge from a more powerful model.
By taking advantage of the proposed idea, for example, we can improve the accuracy of models that necessarily have to be lighter. 
For instance, neural models that must be performed on a mobile device, where the most powerful models cannot be performed due to the low resource's device.

\begin{figure}
    \centering
    \includegraphics[width=.55\textwidth]{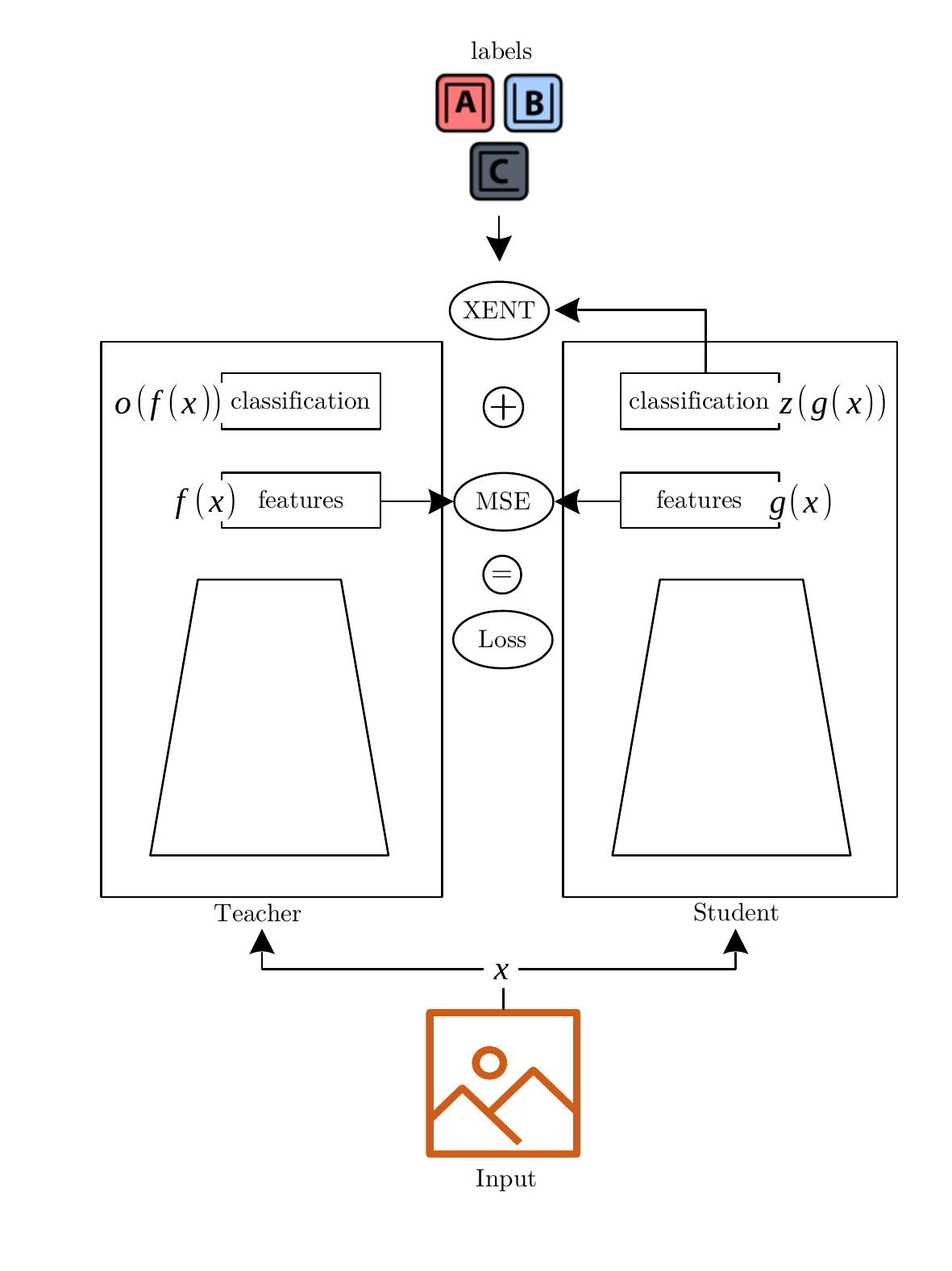}
    \caption{This figure summarizes the proposed TS-Learning model: on the left the teacher, that is the previously trained neural network; on the right, the student that is a weaker model compared to the teacher and is used during the training phase through the MSE loss.}
    \label{fig:proposed_model}
\end{figure}

The solution we propose uses two models: one of the most powerful neural networks trained to solve a certain problem that we call the teacher and a second weaker network that learns to solve the problem with traditional training plus teacher support. This second network is called a student.
As shown in Fig.~\ref{fig:proposed_model}, the proposed model transfers the knowledge of the teacher to that of the student using the second to last layer of the two models and minimizing the Mean Square Error between these two.
In this way, the student can learn a classification problem using the normal cross-entropy loss function and to minimize the error between the knowledge acquired by the teacher and that acquired by his normal learning process.
Thanks to this approach we can also present new random images to the already trained teacher and associate these with the truth label generated by the teacher. 
The student can further improve his performance by taking advantage of the teacher's knowledge through these new labelled examples.

\begin{figure}
    \centering
    \includegraphics[width=.80\textwidth]{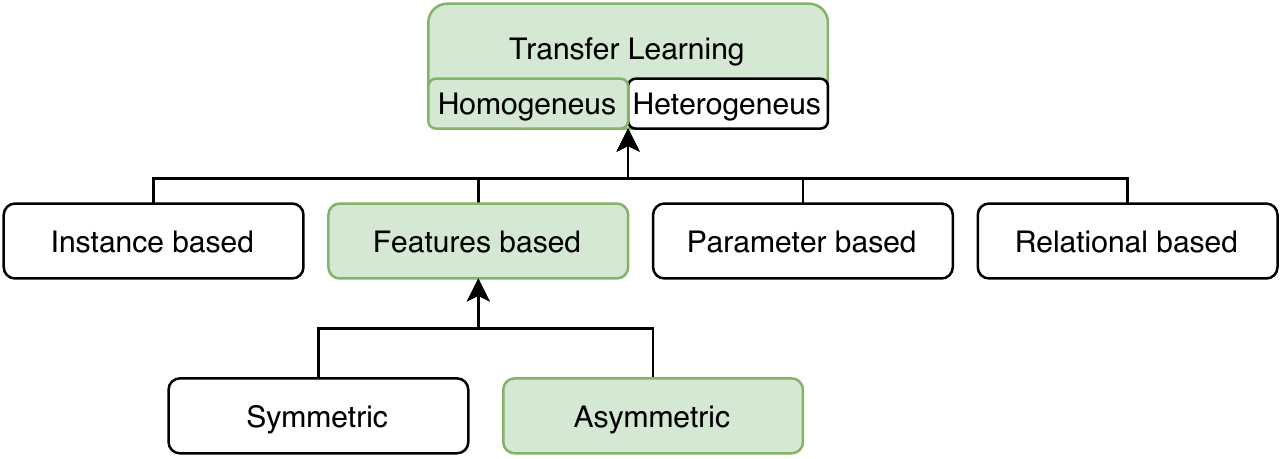}
    \caption{Taxonomy for transfer learning methods. 
    The method proposed in this paper can be defined as a homogeneous, feature-based and asymmetric transfer learning method (green boxes).}
    \label{fig:transfer_learning}
\end{figure}

Following what has been proposed in \cite{pan2009survey,weiss2016survey} on the history and taxonomy of transfer learning, here we try to propose a taxonomic classification of our solution, as proposed in Fig~\ref{fig:transfer_learning}.
Our proposed method can be classified as a homogeneous transfer learning with an asymmetric feature-based approach. This because it is based on the features layer (before the classification layer), and the sizes of the features layers for teacher and student are the same. Furthermore, our method is asymmetric because the student learns from the teacher while the latter no longer learns anything during the student's training.
As previously mentioned, our approach uses data augmentation to improve student outcomes.
In particular, by creating random images, our approach differs from the others because it tries to simulate a teacher who tries to create some simple and not real cases to explain the theory taught by feature transfer.
This generated data are automatically labelled from the already trained teacher model, so our proposed method falls inside the semi-supervised learning category.
In sub-section~\ref{cap:expensivness} we also face up the expensiveness problem, citing all the main methods available in the literature. This because in our solution we are using students that are smaller than the teacher but we know that many other solutions exist.

Collecting enough data to train a deep neural network is very expensive, so the traditional transfer learning approach attempts to start the training process from an available large dataset to pre-train the network and subsequently refine the training with a smaller dataset of the specific domain. This is the traditional approach to transfer learning~\cite{zhuang2019comprehensive,long2017deep}.
Transfer learning can be divided into two categories: \textit{homogeneous} and \textit{heterogeneous transfer}
learning \cite{weiss2016survey}. Homogeneous transfer learning approaches are developed and proposed for handling situations where the domains are of the same feature space. Heterogeneous transfer learning is the scenario where the source and target domains
are represented in different feature spaces. 
The homogeneous transfer learning approach can be split into different sub-categories (see Fig.~\ref{fig:transfer_learning}): \textit{instance‑based},  \textit{feature-based}, \textit{parameter-based} and \textit{relational-based} \cite{pan2009survey}.
Instance-based transfer learning approaches are mainly based on the instance weighting strategy. 
The feature-based approach can be subdivided into other two sub-categories: \textit{symmetric} and \textit{asymmetric}.
Asymmetric approaches transform the source features to match the target ones. In contrast, symmetric approaches attempt to find a common latent feature space and then transform both the source and the target features into a new feature representation.
The parameter-based transfer learning approaches transfer the knowledge at the model/parameter level.
Relational-based transfer learning approaches mainly focus on the problems in relational domains. Such approaches transfer the logical relationship or rules learned in the source domain to the target domain.

\section{Related methods}
Some solutions that attempt to resolve training problems caused by the use of a small dataset are represented by data augmentation techniques.
The data augmentation field can be divided into two classes: basic image manipulation and deep learning manipulation~\cite{shorten2019survey}.
Deep learning based techniques generally use GANs to generate new data.
The basic image manipulation approach instead uses some transformation starting from dataset images, for example, colour space transformation, geometric transformation, random erasing, kernel-based transformation, etc.

Semi-supervised learning \cite{zhai2019s4l} is a machine learning task and method that lies between supervised learning (with completely labelled instances) and unsupervised learning (without any labelled instances). Typically, a semi-supervised method utilizes abundant unlabeled instances combined with a limited number of labelled instances to train a neural network. 
Applying some data augmentation method as automatic label or data generation the classical supervised learning became semi-supervised.

\label{cap:expensivness}
Another important problem is the prediction time. 
In some cases, some deep neural network is too computationally expensive to be used in some devices, so it is typically to use a smaller network. 
Typical situations where these problems can occur are when we try to use these models on mobile devices or real-time devices.
One of the solutions is to include some hardware-based model that use transfer learning \cite{whatmough2019fixynn}. 
In a real-time system, this solution can be considered the best, but not in a general-purpose device as a mobile device. 
So the general problem is not solved up to now, a mobile device needs the simplest models.
A technique used to create simple models is to build a neural architecture to achieve an optimal balance between accuracy and computational power. This point can be reached using the most suitable fixed architecture for a given device as Alexnet \cite{krizhevsky2012imagenet}, GoogLeNet \cite{szegedy2015going} and MobileNet \cite{sandler2018mobilenetv2}.
A small architecture can be obtained also in a not fixed way.
Recently there has been lots of progress in algorithmic architecture exploration included hyperparameters optimization \cite{bergstra2012random} \cite{snoek2012practical} \cite{elsken2018neural} \cite{snoek2015scalable}.
As well as various methods of networks pruning are implemented~\cite{masson2019survey} and connectivity learning \cite{ahmed2017connectivity} \cite{veniat2017learning}.
Also quantization~\cite{wu2016quantized} can be applied to a neural network to improve his computational power.
All this method try to face up the expensiveness problem, but the best computational solution is to have a smaller network.
A transfer learning method similar to that proposed here is the distillation \cite{hinton2015distilling}. 
It tries to transfer the learning between two models using as target truth the response of the pre-trained model, and using as loss function $1/T (z_i - v_i)$ where $T$ is a hyper-parameter named temperature, $z_i$ is the new model prediction using softmax and $v_i$ is the pre-trained model prediction using softmax. 
This method works well if we transfer the learning between an ensemble of the same models, but it does not work well between two different models. 
This solution uses the classification level to transfer knowledge from the teacher to the student, our solution instead is based on the level just before, which is more general.
Another similar method is the variational information distillation \cite{ahn2019variational}. 
It minimizes the cross-entropy as in our proposal while retaining high mutual information with the teacher network. 
The mutual information is computed on more than one layer. 
But its major limitation is that the shape of all of the layers in which the mutual information is computed must be the same. 
Generally, in this method, the teacher network has the same size as the student network, because he aims to transfer learning cross-domain.

\section{Proposed method}
The proposed approach is graphically represented in Fig.~\ref{fig:proposed_model} and its main objective is to obtain a deep neural network capable to obtain better results thanks to a transfer learning process for a specific problem.
The model consists of two actors: the \textit{teacher} (the deep neural network represented on the left in Fig.~\ref{fig:proposed_model}) and the \textit{student} (the deep neural network represented on the right in Fig.~\ref{fig:proposed_model}).
We can call this method Teacher-Student Learning or TS-Learning.
Having fixed a specific problem to be solved, the teacher is one of the deep neural networks that best solves the problem, while the student is a model with a lower number of parameters which, during the training process, exploits the knowledge acquired by the teacher to produce better results.
A constraint for these two deep neural networks is that they must have the layer located just before the output layer with identical sizes.
Also, the last layers of the two networks have the same shape because they are trained on the same data set.

Like in Hinton \textit{et al.}~\cite{hinton2015distilling} and Ahn \textit{et al.}~\cite{ahn2019variational} we use a teacher that is a pre-trained model to train the student, but teacher and students are different and have the constraint that the two layers before output layers must have the same size. In particular, the teacher model in \cite{hinton2015distilling} is an ensemble of classifiers, and in \cite{ahn2019variational} is the same as the student.
The model proposed in \cite{hinton2015distilling}  tries to train again the same model using the classification layer obtained from an ensemble of classifiers.  
In our TS-Learning, we train the last layer of a student using the cross-entropy loss function and transferring the learning only between the second-last layers through another loss function based on Mean Squared Error (MSE) between second-last layers of teacher and student. 
In \cite{ahn2019variational} the authors use the same loss function for the last layers but minimize also mutual information from all other layers because, in their model, teachers and students are the same models, but the teacher is trained on another dataset.
As transfer loss function we use an MSE loss instead in \cite{hinton2015distilling} the authors use a simple subtraction of the two softmax functions used for teacher and student.
In~\cite{ahn2019variational} the authors use a variational method and in particular, they use a kl-divergence into the loss function. In our proposed approach we do no use it, but in future, it is possible to introduce it also in our TS-Learning.

Going into detail, the proposed TS-Learning is trained through two main steps: during the first step, the teacher is trained and after that, we start to train the student summing the two proposed loss functions as graphically represented in Fig.~\ref{fig:proposed_model}.
Let $\mathcal{D}$ be a dataset containing the set of pairs $(x_i, y_i)$, where $x_i$ is an input image and $y_i$ is its ground-truth label.
In this work, the teacher and student are two different CNNs for images classification, but this is not a strong constraint.
A strong constraint that must be respected is related to the size of the two output layer (both networks must be used for the same classification problem and therefore with the same set of classes $c$) and to the size $d$ of the two feature layers. Following the graphic representation of Fig.~\ref{fig:proposed_model} we want that the constraints of Eq.~\ref{eq:constraints} are respected.
\begin{equation}
  f(x), g(x) \in \mathbb{R}^d \land o(f(x)), z(g(x)) \in \mathbb{R}^c
  \label{eq:constraints}
\end{equation}
where $f(x)$ is the  features extractor of the Teacher, and $g(x)$ is the  Student  feature  extractor.

To train the teacher we use a \textit{softmax cross entropy} loss function.
Softmax converts the output logits, $a_i$, computed for each class into a probability $p_i$, by comparing $a_i$ with the other logits.
Softmax function takes an $n$-dimensional vector of real numbers and transforms it into a vector of real number in the range $[0,1]$ which add up to 1. 
\begin{equation}
  p_i = \frac{e^{a_i}}{\sum_j e^{a_j}}
\end{equation}
Cross entropy indicates the distance between what the model believes the output distribution should be, and what the original distribution really is ($y_i$). 
It is defined as
\begin{equation}
\mathcal{L}_{xent}(y, p)=-\sum_{i} y_{i} \log \left(p_{i}\right)
\end{equation}
Since in the training process we update the weights at each batch of size $m$, then the loss function for the teacher becomes the following
\begin{equation}
  \mathcal{L}_{xent} = - \sum_{i=1}^{m} y_i \cdot log \frac{e^{a_{i}}}{\sum_{j=1}^{n} e^{a_j}}
  \label{eq:cross-entropy-with-softmax_loss}
\end{equation}
where $n$ is the number of classes. 

After training the teacher, we can train the student to take advantage of the teacher's knowledge during the training process.
To train the student network we use the sum of two-loss functions: the softmax cross-entropy defined in Eq.~\ref{eq:cross-entropy-with-softmax_loss} plus the MSE loss, as defined in the following equation
\begin{equation}
    \mathcal{L} = {\lambda}_{mse} \cdot \mathcal{L}_{mse} + {\lambda}_{xent} \cdot \mathcal{L}_{xent}
\label{eq:total_loss}
\end{equation}
where ${\lambda}_{mse}$ and ${\lambda}_{xent}$ are two scalar used to balance the two loss functions.
The $\mathcal{L}_{mse}$ loss is the Mean Squared Error loss defined in the following Eq. \ref{eq:mse} and normally used for regression problems.
So we can say that the proposed method is a classification method that also tries to regress the function $f(x)$ learned by the teacher.
\begin{equation}
    \mathcal{L}_{mse} = \frac{1}{md} \sum_{i=1}^{m}  {\sum_{j=1}^{d} (f_{j}(x_i) - g_{j}(x_i))^2 }
\label{eq:mse}
\end{equation}

Our proposal is based on the simple conjecture that there exists a function $g(x)$ which can approximate the function $f(x)$, and $g(x)$ is computationally less expensive than $f(x)$.
In particular we want that after training the student network, $g(x) \approx f(x)$.
In our proposal we try to approximate $f(x)$ with $g(x)$ and at the same time we try to approximate the ideal function with $z(g(x))$. 
In this way, we can improve the accuracy of the student network using $z(g(x))$.


To further improve the student's generalization ability, we have introduced a data augmentation strategy in the student network training process.
In practical implementation, we use a generic random generator able to represent a sequence of real numbers and to build a two-dimensional array in a range $[0,1]$.
We use a two-dimensional array like random images and generate a group of different random images for each batch size as $|batch\ size|=|batch\ size| * N$ where N is an integer chosen that represents the number of random images to be created.
In literature exists a wide range of algorithms to get these deterministic sequences of values and a classical example of a generator can be the linear congruential generator.
Given $a, c, m$ be variables, the generator assume the follow form:
$X_{n+1} = (aX_{n} + c) \ mod \ m$
where $m$ is some chosen modulus and $X_0$ is an initial number well-known as seed.
A generator is cyclic and after a finite number of calls, it again returns the same sequence generated for the first time.
This period is in the order of $2^{31}-1$ or the recent algorithm more robust is $2^{19937}-1$, which is more than sufficient to generate $N$ random different images for our data augmentation approach. So, for each training batch a fixed number of random generated $\hat{x}_i$ images are used by the teacher network to obtain the predicted $\hat{y}_i$ label from it and thus used to improve the student's generalization ability.

\section{Datasets}
To demonstrate the quality of the TS-Learning model qualitatively and quantitatively, we used the following well-known datasets.

\textbf{Cifar10} \cite{cifar} dataset consist of 60000 images divided in 10 classes (6000 per classes) with a training size and test size of 50000 and 10000 respectively.
Each input sample is a $32\times 32$ colour images with a low resolution. 
The 10 classes are airplanes, cars, birds, cats, deer, dogs, frogs, horses, ships, and trucks. 

\begin{figure}
    \centering
    \includegraphics[width=.90\textwidth]{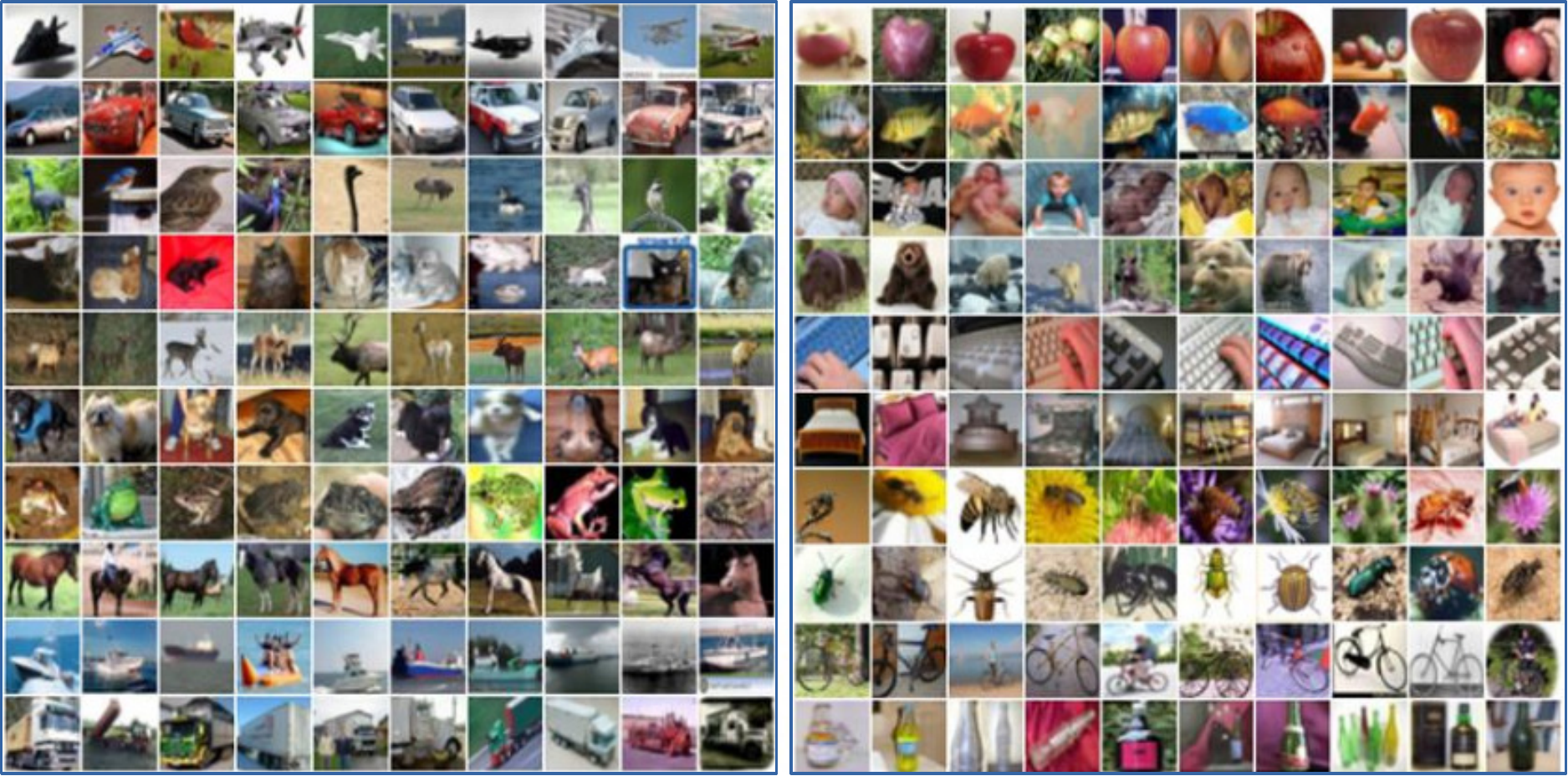}
    \caption{On the left some sample images for each of the 10 classes of CIFAR-10 (one class for each row). On the right 10 classes randomly selected from the set of 100 classes of CIFAR-100.}
    \label{fig:cifar10-100}
\end{figure}

\textbf{Cifar100} \cite{cifar} dataset consist of 60000 images divided in 100 classes (600 per classes) with a training size and test size of 50000 and 10000 respectively.
Each input sample is a $32\times 32$ colour images with a low resolution.

In Fig.~\ref{fig:cifar10-100} some representative examples of the two datasets, extracted from a subset of classes.

\textbf{Cub200 2011}~\cite{WahCUB_200_2011} is a dataset that we use to face up the fine grained classification problem. 
It contains $11788$ images of $200$ species of birds split in $5994$ and  $5794$ images for train and test respectively.
Each image contains $15$ part locations,  $312$ binary attributes and  $1$ bounding box containing the bird within the image, but we used only input images and their labels.

\begin{figure}
    \centering
    \includegraphics[width=.90\textwidth]{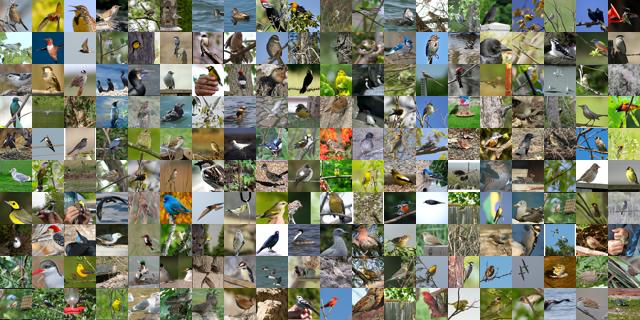}
    \caption{A sample image for each of the 200 classes of the CUB200-2011 dataset.}
    \label{fig:cub200}
\end{figure}

\section{Experiments}
We now present our experimental results.
We first compare a simple deep model published in the literature trained with the classical approach and trained with our TS-Learning, to try to understand if the same model thanks to the contribution of a Teacher can improve its accuracy.
As a second experiment, we want to compare different Students who use the same Teacher, to understand in which situation a Student can take full advantage of the Teacher.
In the last experiment, we want to test our solution on a very difficult fine-grained classification problem, to understand if a standard Student can learn the solution better in such a difficult context thanks to the help of a very good Teacher.
In all our experiments we used ${\lambda}_{mse} = {\lambda}_{xent} = 1$ and Adam~\cite{kingma2014adam} optimizer with learning rate value as $0.001$, except for the Cifar10 Alexnet~\cite{krizhevsky2012imagenet} and ResNet18~\cite{He2015,targ2016resnet} that have different learning rate respectively of $0.0001$ and $0.01$ in order to perform better results.
The learning rate decreases every 50 epochs during training by a factor 0.1.

In the first experiment, we compare our solution with the simple model proposed in~\cite{wu2015max}, using the Cifar10 dataset.
As described in~\cite{wu2015max}, it is a slightly modified version of the original LeNet5 network.
We used the same model as Student by changing the dimension $d = 2048$ of its $g(x)$ layer to adapt it to the dimension of the Teacher's layer $f(x)$.
The Teacher is an already trained ResNet56~\cite{He2015,targ2016resnet} whose size of its $f(x)$ layer  is $d = 64$.
Since we have reduced the size of the layer quite a bit, we expect a loss in accuracy by comparing this model modified by us with the original one.
Analyzing the results reported in Tab.~\ref{tab:lenetpaper} it can be seen that the accuracy of the LeNet5 modified by us is 78\% while that published in~\cite{wu2015max} is slightly higher.
Using this model as Student and ResNet56 as Teacher, it can be seen that there is an increase in accuracy concerning the published model.
By adding the data augmentation strategy, as reported in Tab.~\ref{tab:lenetpaper}, we obtain a minimal improvement.
In this case, we added about $10\cdot |c|$ random images to each batch of size 128, where $|c|$ is the number of classes.
\begin{table}
    \caption{Accuracy obtained with the proposed TS-Learning model using the Cifar10 dataset, compared with the model used in~\cite{wu2015max}. The Student is a modified version of a LeNet5 CNN, while the Teacher is a ResNet56 CNN. }
    \label{tab:lenetpaper}
    \begin{center}
        \begin{tabular}{cccc} 
            \hline
            Paper~\cite{wu2015max}~~ & Student~~ & Student+Teacher~~ & Student+Teacher+Aug.\\
            \hline
            0.804 & 0.78 & 0.8136 & 0.8149\\
            \hline
        \end{tabular}
    \end{center}
\end{table}
%

\begin{figure}
    \centering
    \includegraphics[width=.75\textwidth]{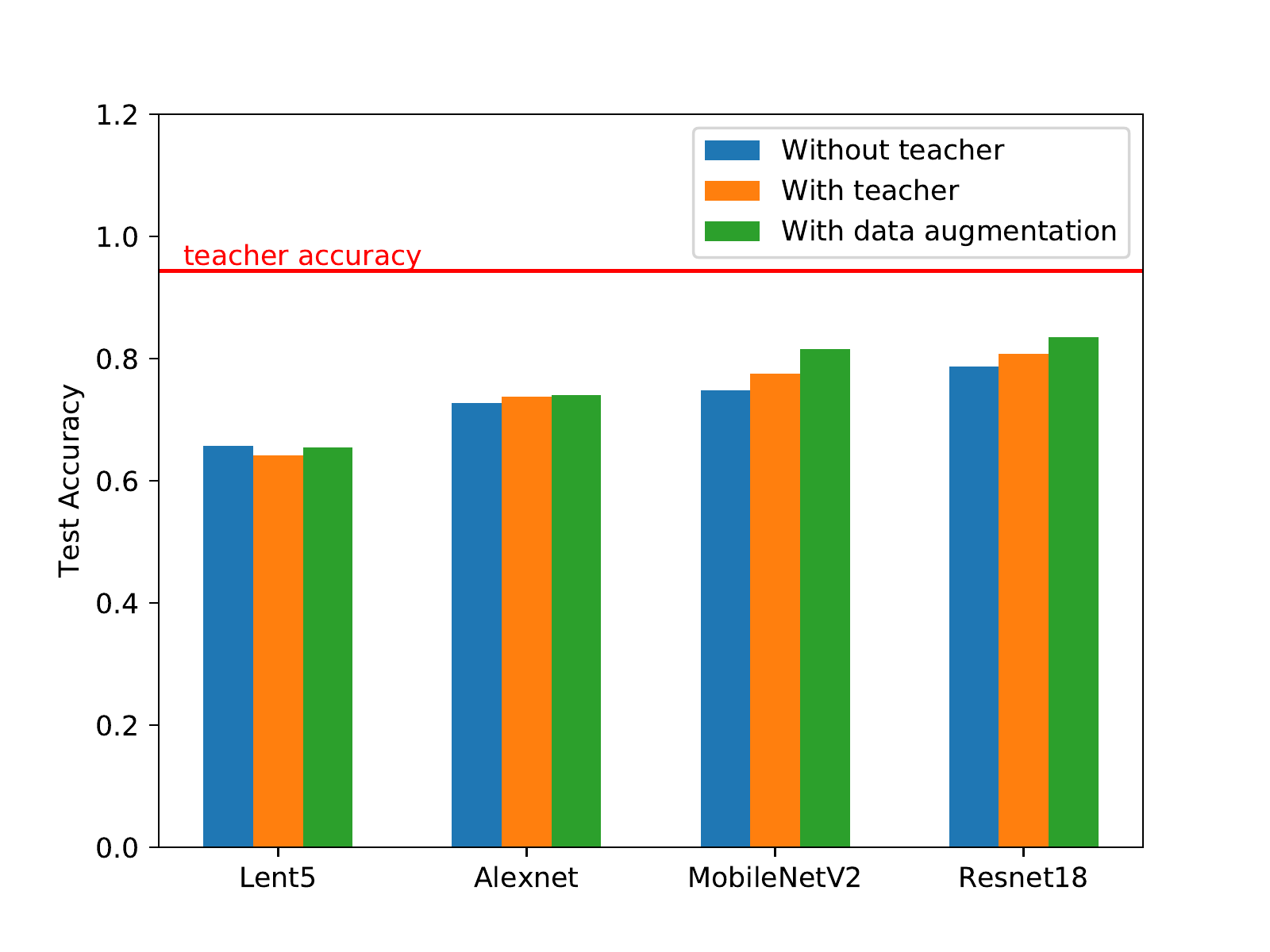}
    \caption{Test accuracy obtained with four different Students using the Cifar10 dataset.
    For each Student we compare the test accuracy by training the model without Teacher, with the Teacher and with the Teacher plus the data augmentation.}
    \label{fig:c10_barplot}
\end{figure}

\begin{figure}
    \centering
    \includegraphics[width=.80\textwidth]{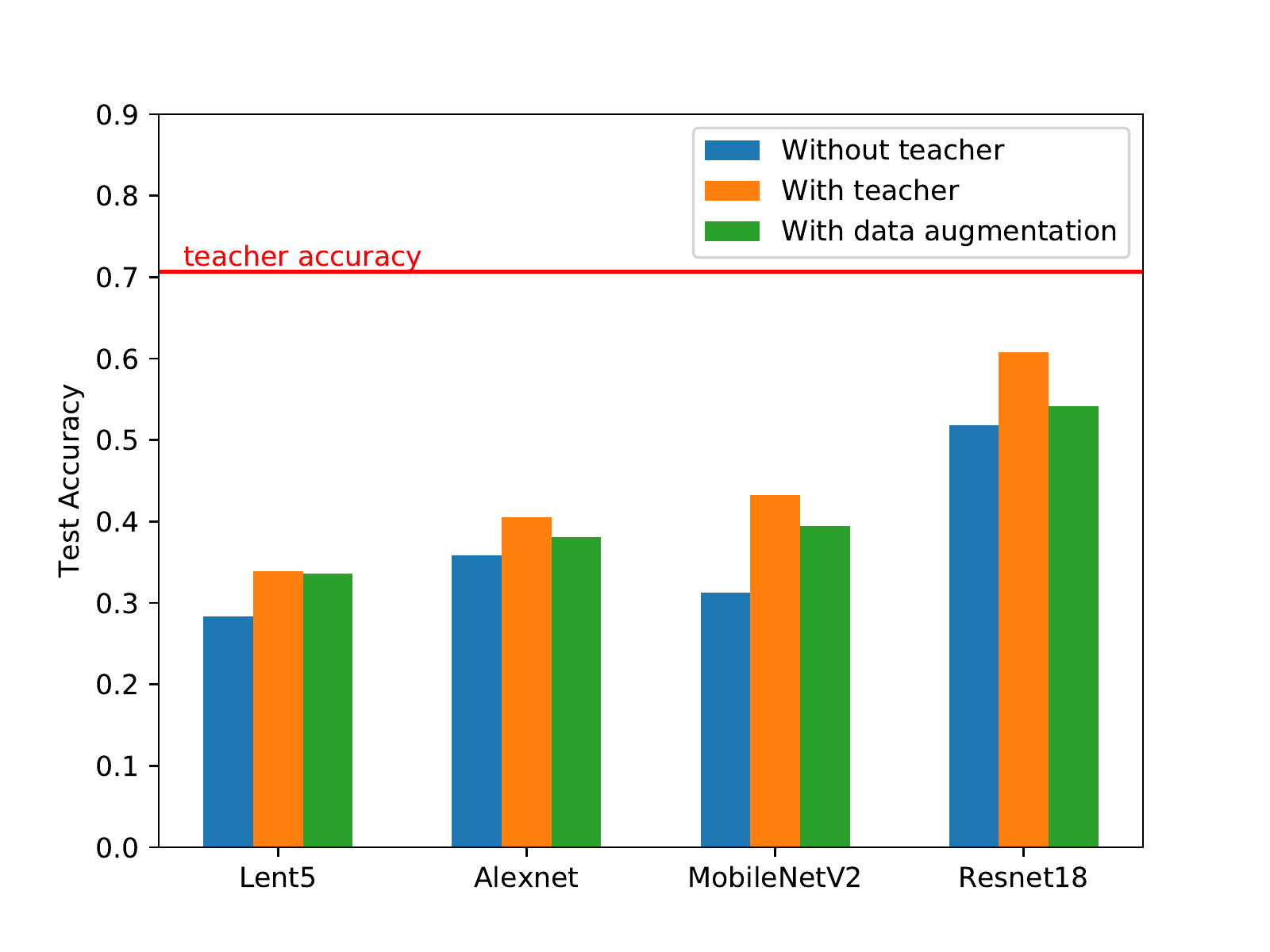}
    \caption{Test accuracy obtained with four different Students using the Cifar100 dataset.
    For each Student we compare the test accuracy by training the model without Teacher, with the Teacher and with the Teacher plus the data augmentation.}
    \label{fig:c100_barplot}
\end{figure}

\begin{figure}
    \centering
    \includegraphics[width=.80\textwidth]{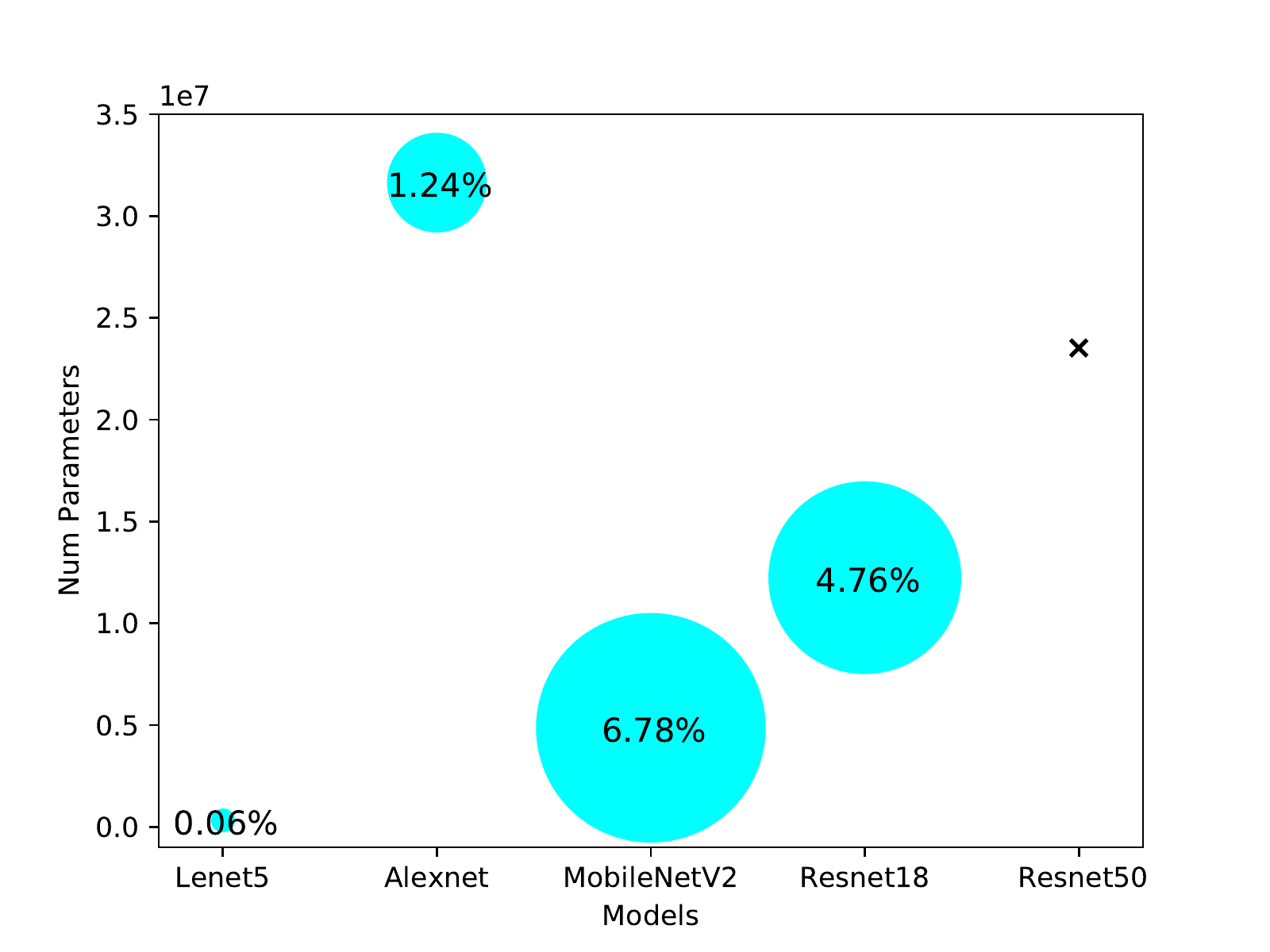}
    \caption{Accuracy increment ($acc_{stu + tea} - acc_{stu}$) obtained on Cifar10 using four different networks as Students, ordered from left to right by discriminating ability. 
    We report on the Y-axis the number of parameters contained in each Student.
    The black cross represents the Teacher used.}
    \label{fig:c10_paramiters}
\end{figure}

\begin{figure}
    \centering
    \includegraphics[width=.80\textwidth]{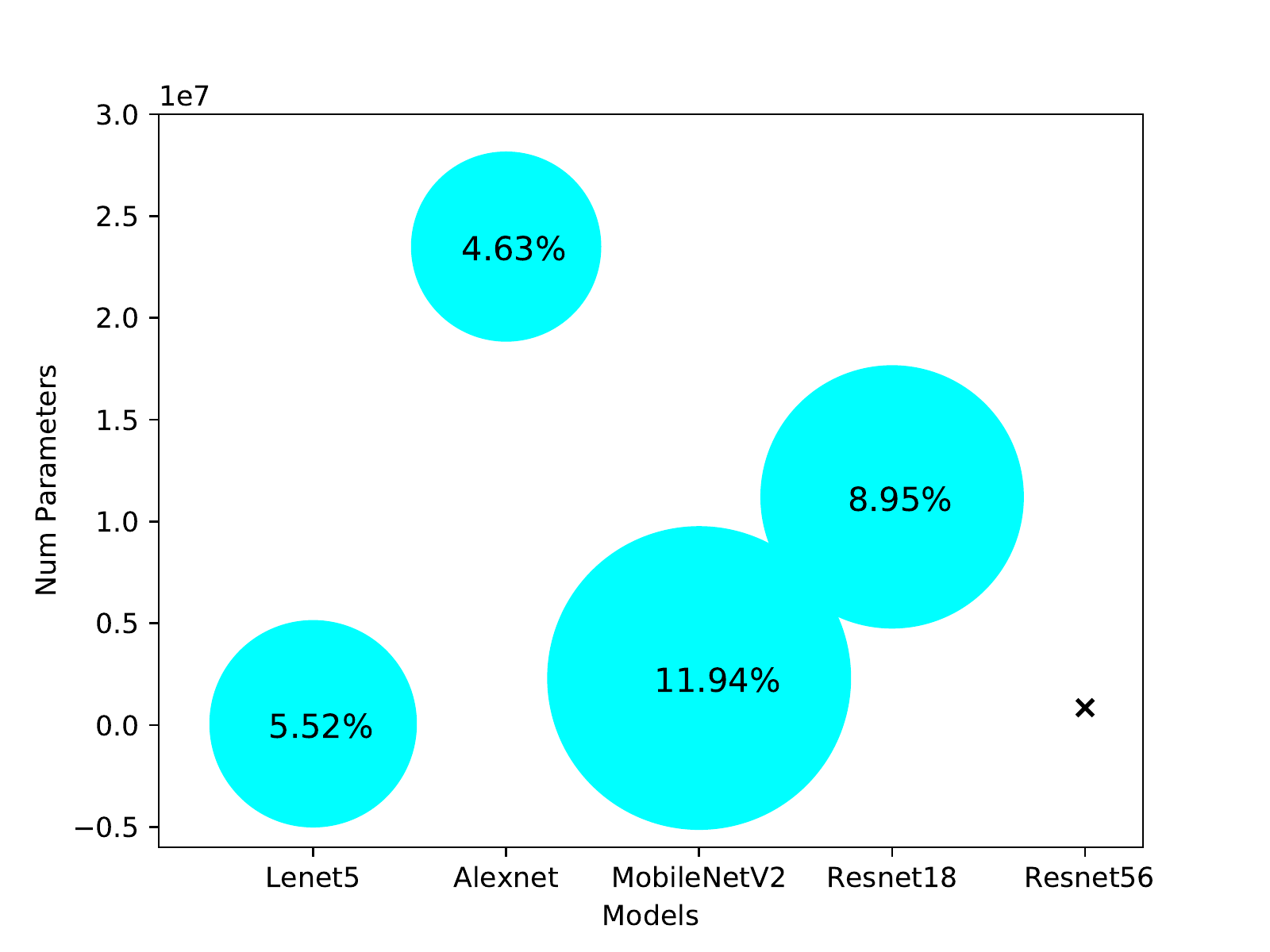}
    \caption{Accuracy increment ($acc_{stu + tea} - acc_{stu}$) obtained on Cifar100 using four different networks as Students, ordered from left to right by discriminating ability. 
    We report on the Y-axis the number of parameters contained in each Student.
    The black cross represents the Teacher used.}
    \label{fig:c100_paramiters}
\end{figure}

\begin{figure}
    \centering
    \includegraphics[width=.48\textwidth]{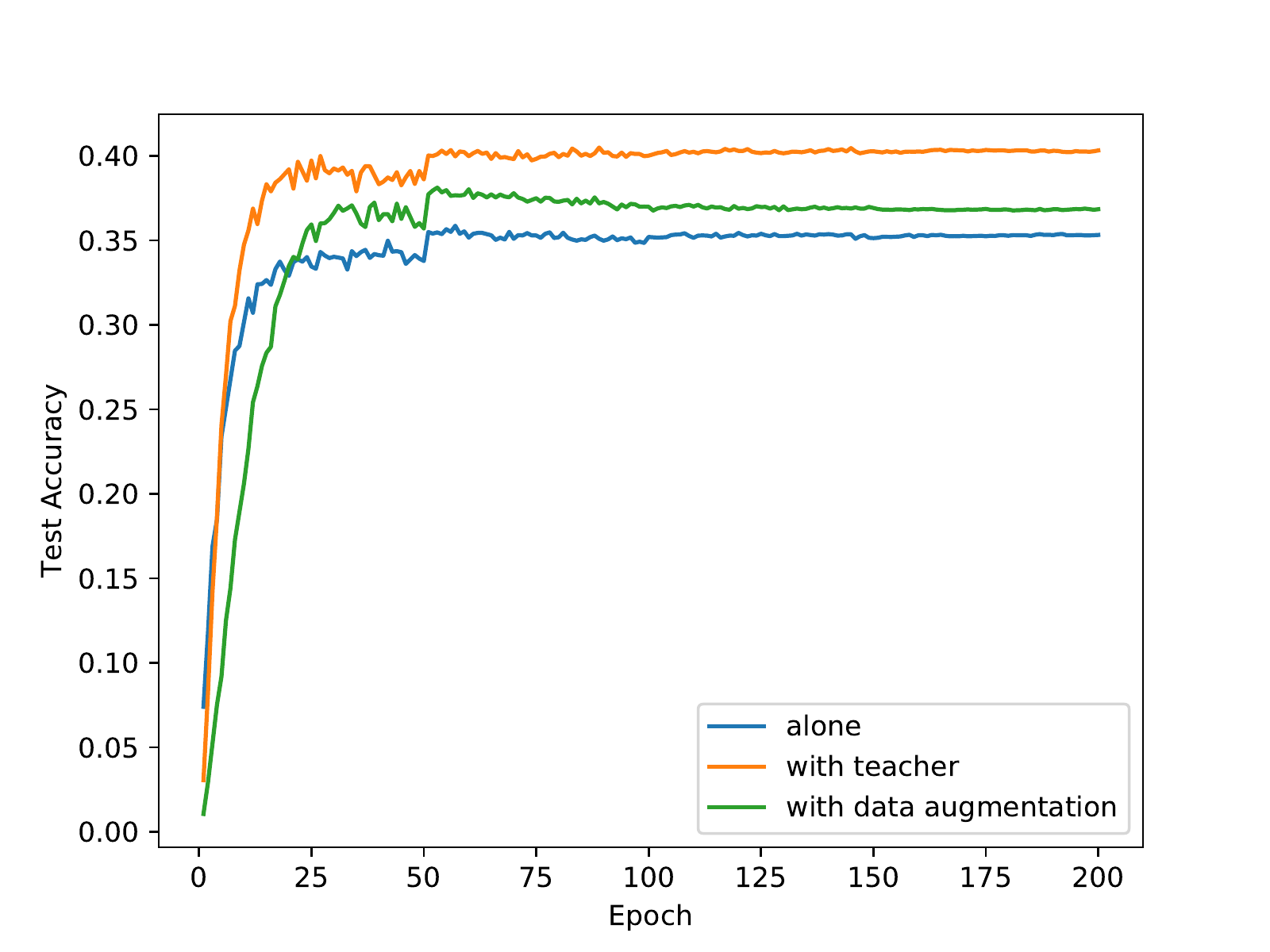}
    \includegraphics[width=.48\textwidth]{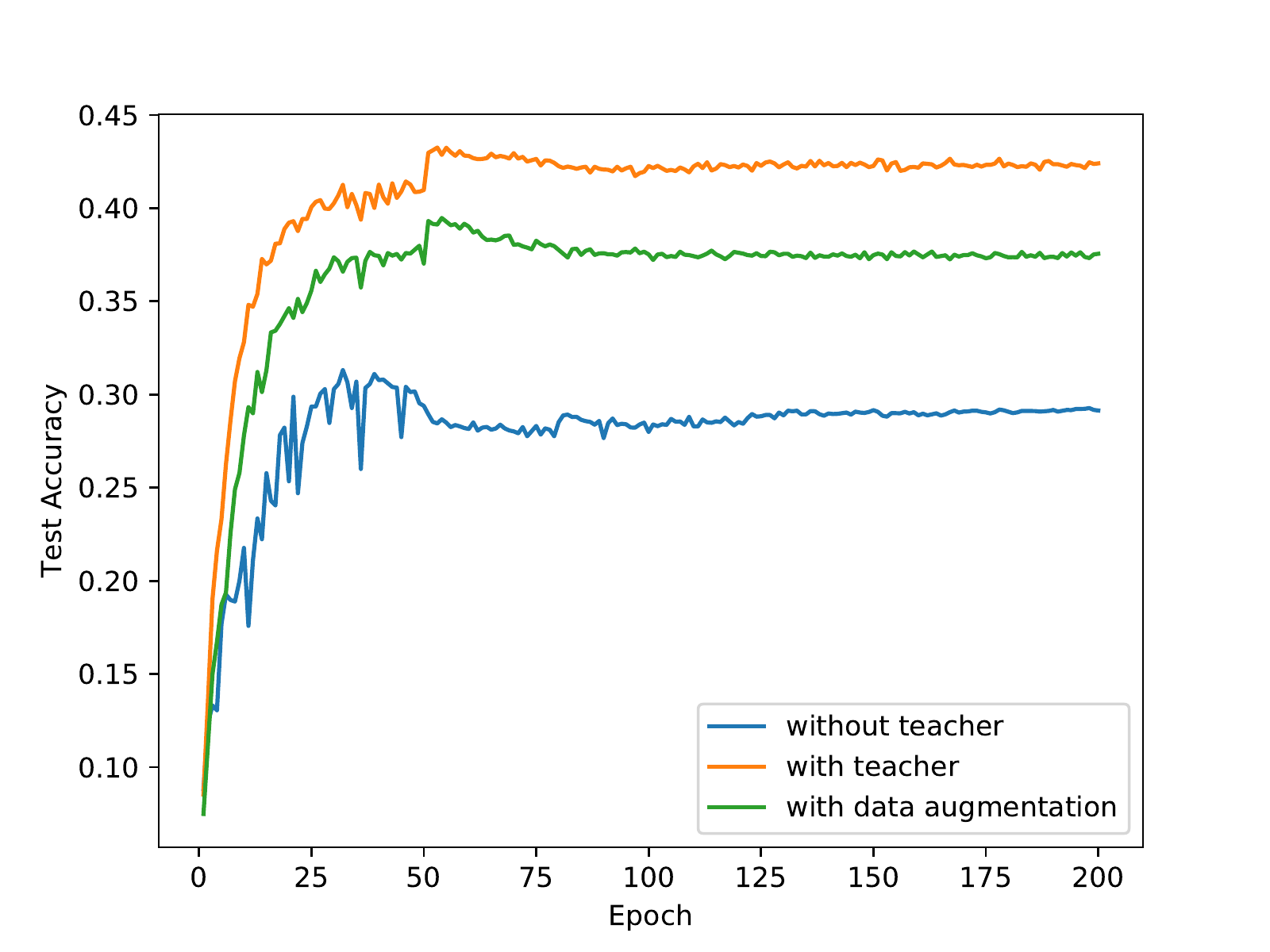}
    \caption{Two typical behaviors that are obtained on the Cifar100 test set during the training of the Student only, the Student plus the Teacher and the Student plus Teacher plus Augmentation. On the left the student is an AlexNet while on the right is a MobileNetV2.  For both plots, the Teacher is a ResNet56.}
    \label{fig:alexnet_c100}
\end{figure}

In the second experiment, we compare different CNNs used as Students.
Different CNNs produce different accuracies on some benchmark datasets~\cite{canziani2016analysis}. 
In this experiment we use the following networks which in the order produce increasing accuracy: LeNet5~\cite{lecun2015lenet}, AlexNet~\cite{krizhevsky2012imagenet}, MovileNetV2~\cite{sandler2018mobilenetv2} and ResNet18~\cite{He2015,targ2016resnet}.
As Teacher model, we used a pre-trained ResNet50~\cite{He2015,targ2016resnet} for Cifar10 and ResNet56~\cite{He2015,targ2016resnet} for Cifar100. Since the Teacher has a layer $g(x)$ of a fixed size (2048 for the ResNet50 and 64 for the ResNet56) then we have modified the layers $g(x)$ of all the Students so that they match the same size $d$ of the Teacher.
In this experiment, we used the Cifar10 and Cifar100 datasets.
The batch size is 128 with a data augmentation of 178 (empirically set) to have a real batch size of 306 images. 
In Tab.~\ref{tab:tableComparison} and Figs.~\ref{fig:c10_barplot}, \ref{fig:c100_barplot}, \ref{fig:alexnet_c100} we report the maximum accuracy obtained after 200 training epochs by comparing the result obtained by the single model used as a Student with the results of the same model trained using the Teacher and the Teacher plus the data augmentation.
Figs.~\ref{fig:c10_paramiters}, \ref{fig:c100_paramiters} show the increases in accuracy ($acc_{stu+tea} - acc_{stu}$) obtained by individual Students (compared to the same Student without Teacher) on the two Cifar datasets.
Thanks to these experiments we can conclude that using the Teacher always leads to a Student with higher test accuracy. Only the LeNet5 with the Cifar10 cannot improve because it is not powerful enough to imitate (or learn) from its Teacher (ResNet50).
On the group of experiments done on the Cifar100 dataset, we can say that the data augmentation strategy does not lead to an improvement compared to the use of only the Student plus Teacher because a set of 178 random images per batch is not enough to have a balanced set divided over 100 classes.
In general, with or without data augmentation we always get better results by comparing ourselves with the Student without Teacher.
\begin{table}
    \caption{Comparison of all the test accuracy obtained with the Cifar10 and Cifar100 datasets, after 200 epochs, using four different Students and two different Teachers.}
    \label{tab:tableComparison}
    \begin{center}
        \begin{tabular}{llccc} 
            \hline
            Name & Student~~ & Student+Teacher~~ & Student+Teacher+Aug.\\
            \hline
            Cifar10 & & & \\
            \hline
            Lenet5 & 0.6543 & 0.6412 & \textbf{0.6549}\\
            AlexNet & 0.7275 & 0.7374 & \textbf{0.7399}\\
            MobileNetV2~ & 0.7482 & 0.7753 & \textbf{0.816}\\
            Resnet18 & 0.7876 & 0.8081 & \textbf{0.8352}\\
            \hline
            Cifar100 & & & \\
            \hline
            Lenet5 & 0.2834 & \textbf{0.3386} & 0.3365\\
            AlexNet & 0.3586 & \textbf{0.4049} & 0.3812\\
            MobileNetV2 & 0.3131 & \textbf{0.4325} & 0.3947\\
            Resnet18 & 0.5182 & \textbf{0.6077} & 0.5418\\
            \hline
        \end{tabular}
    \end{center}
\end{table}
%

In the last experiment, to validate the goodness of our TS-Learning model, we train a MobileNetV2~\cite{sandler2018mobilenetv2} on the CUB200 dataset using a pre-trained~\cite{nawaz2019these} NTS-Net~\cite{yang2018learning} model as Teacher. 
The layer $g(x)$ of the MobileNetV2 Student is adapted to the size of $10240$ of its Teacher.
We use a batch size of $10$ real image.
In these experiments, we used input images with size $448\times 448$ obtained scaling the original image to $600\times 600$ and extracting a centered crop of size $448\times448$.
To perform the best results the experiments without the Teacher have a learning rate of $0.0001$ and it does not change during the epochs, instead of the TS-Learning experiment have a learning rate of $0.001$ and after $100$ epochs it is multiplied by $0.1$.
In the Table~\ref{tab:bird} we can see the Student's accuracy without and with a Teacher.
\begin{table}
    \caption{Test accuracy of a MobileNetV2 used as a Student with an NTS-Net used as a Teacher after 200 training epochs. The dataset used is CUB200.}
    \label{tab:bird}
    \begin{center}
        \begin{tabular}{lcc} 
            \hline
            Name & Without teacher & With teacher\\
            \hline
            MobileNetV2 & 0.425 & \textbf{0.492}\\
            \hline
        \end{tabular}
    \end{center}
\end{table}
The  NTS-Net \cite{yang2018learning} is a model used for fine grained classification, concatenating the features extracted from the whole image and four crops using attention mechanisms. 
Using our solution, the Student network learns to extract the same features so he learns implicit attention. 
As we can see in Table~\ref{tab:bird}, the Student network can improve his accuracy tanks to our TS-Learning strategy.

\section{Conclusion and future works}
We described a feature transfer method to transfer knowledge between two different models adding a second loss function in addiction to standard cross-entropy. 
We proposed also a data augmentation strategy with random image generation to improve the test accuracy. 
We show that the proposed method is particularly suitable for mobile applications because it is possible to improve the accuracy of the simplest models using the knowledge learned by more complex models.

For the Cifar10 dataset, our proposed method improves the accuracy in every tested situation.
On Cifar100 dataset, we can say that the data augmentation strategy does not lead to improvement because the set of random images added per batch is not enough for a 100 classes problem. 
In general, with or without data augmentation we always get better results in all the experiments conducted, so we can conclude that using the Teacher always leads to a Student with higher test accuracy.

A limitation of our TS-Learning approach is the size of the feature layer: it must be the same in the Student and Teacher networks. 
A future improvement could use an auto-encoder to translate the information from a shape to another to leave the Student features layer with the original dimensions.
Another improvement to be made could solve the problem of data augmentation in the presence of a large number of classes, as highlighted in the experiment done on the Cifar100 dataset. In this case, we will look for a solution to balance the number of random examples for each class.

Thanks to the proposed approach, in this work we have experimented with a new data augmentation method with random generated images, that we can imagine as the exercises that the Teacher leaves to the Student to improve his knowledge of a problem. 
Finally, the proposed TS-Learning solution improves the results of weak neural networks to be used on devices with few hardware resources and opens new research directions on transfer learning.

\clearpage
%
%
\bibliographystyle{splncs04}
\bibliography{biblio.bib}

\end{document}